\documentclass[letterpaper, 10 pt, conference]{ieeeconf}  

\IEEEoverridecommandlockouts                              

\usepackage{url}
\usepackage{caption}
\usepackage{subcaption}
\usepackage{graphicx}
\usepackage{multirow}
\usepackage{enumerate}
\usepackage{amsmath}
\usepackage{etoolbox}
\usepackage{placeins}
\usepackage{hyperref}
\usepackage{setspace} 
\usepackage{romannum}
\usepackage{wasysym}

\usepackage{bbding}
\usepackage{pifont}
\usepackage{amssymb}
\usepackage{xcolor}
\usepackage{array}
\usepackage{cite}
\usepackage{algorithm}
\usepackage{algpseudocode} 
\usepackage{textcomp}
\usepackage{stfloats}\usepackage{algpseudocode}
\usepackage{hhline}

\begin{document}

\title{\LARGE \
Learning to Place Unseen Objects Stably using a Large-scale Simulation
}

\author{Sangjun Noh$^{*}$, Raeyoung Kang$^{*}$, Taewon Kim$^{*}$, Seunghyeok Back, Seongho Bak, Kyoobin Lee†
\thanks{\text{*} Equally contributed.}
\thanks{All authors are with the School of Integrated Technology, Gwangju Institute of Science and Technology, Cheomdan-gwagiro 123, Buk-gu, Gwangju 61005, Republic of Korea. 
† Corresponding author: Kyoobin Lee {\tt\small kyoobinlee@gist.ac.kr}}%
}

\maketitle


\begin{abstract}
Object placement is a fundamental task for robots, yet it remains challenging for partially observed objects. Existing methods for object placement have limitations, such as the requirement for a complete 3D model of the object or the inability to handle complex shapes and novel objects that restrict the applicability of robots in the real world. Herein, we focus on addressing the \textbf{U}nseen \textbf{O}bject \textbf{P}lacement (\textbf{UOP}) problem. We tackled the UOP problem using two methods: (1) UOP-Sim, a large-scale dataset to accommodate various shapes and novel objects, and (2) UOP-Net, a point cloud segmentation-based approach that directly detects the most stable plane from partial point clouds. Our UOP approach enables robots to place objects stably, even when the object's shape and properties are not fully known, thus providing a promising solution for object placement in various environments. We verify our approach through simulation and real-world robot experiments, demonstrating state-of-the-art performance for placing single-view and partial objects. Robot demos, codes, and dataset are available at \href{https://gistailab.github.io/uop/}{https://gistailab.github.io/uop/}.

\end{abstract}

\section{Introduction} 
Robots need to have the ability to manipulate unseen objects to operate effectively in various environments, which are common in manufacturing, construction, and household tasks. While deep learning has progressed to recognize and handle unseen objects, most of the current research focuses on identifying\cite{xie2021unseen, back2022unseen} or grasping\cite{mahler2017dex, mousavian20196} them. However, it is important to note that when a robot picks up an object from a cluttered container or receives it from a human, the robot must be able to place the object stably. Thus, this study addresses the \textbf{U}nseen \textbf{O}bject \textbf{P}lacement (\textbf{UOP}) problem, which involves stably placing novel objects in a real-world environment. 

\begin{figure}[ht]
    \centering
        \begin{subfigure}[t]{\columnwidth}
            \centering
            \includegraphics[width=\textwidth]{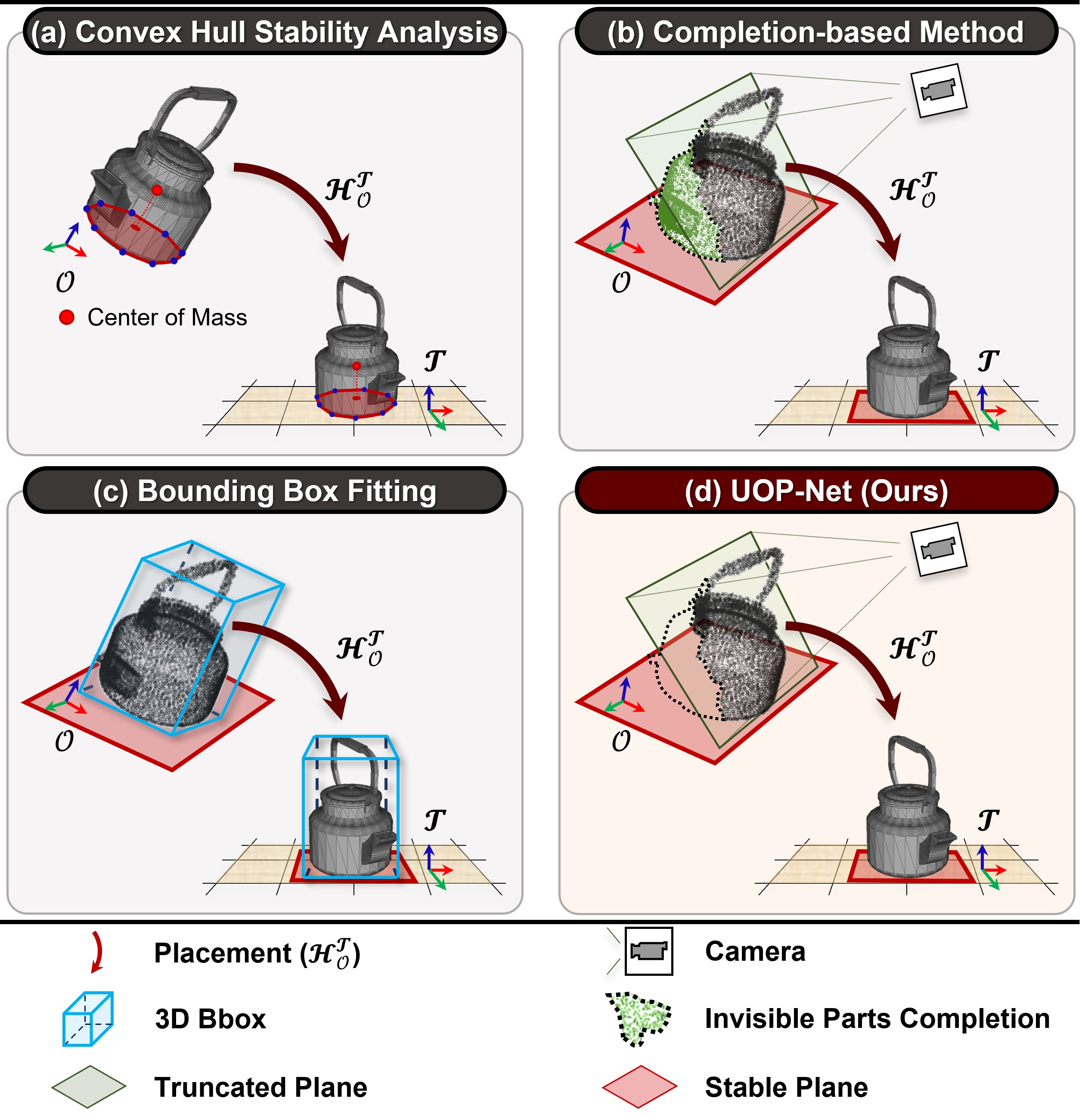}
        \end{subfigure}
        
    \caption{\textbf{Comparison of UOP-Net (Ours) and previous methods.} Previous studies for object placement used (a) full-shape object models\cite{haustein2019object, trimesh, hagelskjaer2019using}, (b) completion modules\cite{gualtieri2021robotic}, or (c) fitted primitive shapes\cite{fischler1981random, Zhou2018}. In contrast, (d) the proposed UOP-Net directly detects stable planes for unseen objects from partial observations.
}    
    \label{fig:task_comparison}
\end{figure}


Conventional approaches\cite{haustein2019object, trimesh, hagelskjaer2019using} for stable object placement require full 3D models and analytical calculations. These methods involve sampling stable planes after calculating the center of mass of the object, which is not feasible for all real-world objects that may be encountered. One approach\cite{gualtieri2021robotic} combines analytical methods with a 3D object completion model that can reconstruct the full shape of an object from raw perception data. However, using this approach is difficult since the predicted point cloud may not be precise, resulting inaccuracies in determining stable planes. Our UOP method addresses these limitations by directly detecting stable planes of unseen objects from single views and partial point clouds, thus eliminating the need for a full 3D object model. This enables the robot to stably place the object even when the shape and properties of the object are not fully known.

In this paper, we propose a method for UOP that detects stable planes from complex shapes and novel objects. To achieve this, we generated a large-scale synthetic dataset called UOP-Sim, which contains various 3D objects and annotations of stable planes generated using a physics simulator. Unlike previous approaches \cite{jiang2012learninga, jiang2012learningb} that rely on heuristics to label the preferred placement configurations, we automatically annotate all feasible planes that can support stable object poses. Our dataset includes 17.4K objects and a total of 69K annotations. We propose a point cloud instance segmentation based network referred to as UOP-Net that predicts stable planes from partial point cloud and train it using only the UOP-Sim dataset. We compare the performance of our approach with three baselines and learning-based methods. We demonstrate that it achieves state-of-the-art (SOTA) performance in both the simulation and real-world experiments without any fine-tuning on real-world data.

The main contributions of this study are as follows:
\begin{itemize}    
    \item{We propose a new task called UOP to place an unseen object stably from single views and partial point clouds.}
    
    \item{We provide a public, large-scale 3D synthetic dataset 
    called UOP-Sim that contains a total of 69,027 annotations of stable planes for 17,408 different objects.}
 
    \item{We introduce a point cloud instance segmentation network named UOP-Net that predicts stable planes for partially observed unseen objects.}

    \item{We compare the performance of our approach with previous object placement methods and confirm that our method outperforms the SOTA methods without any fine-tuning in real-world environments.}

\end{itemize}

\section{Related Works}
\noindent\textbf{Stable object placement.} Previous studies\cite{tournassoud1987regrasping, wan2019regrasp,lertkultanon2018certified, haustein2019object} demonstrated that robots can stably place an object with known geometrical properties by analyzing the convex hull and sampling stable planes for the object. However, this approach requires precise object priors (e.g., CAD, mass), and it may not be available in real-world scenarios with partial observations (e.g., from an RGB-D camera). Several researchers attempted to address this limitation with deep learning-based completion methods that predict the invisible part of an object\cite{gualtieri2021robotic}; however, these approaches have limitations in generating the precise shapes of unseen objects. Our UOP method addresses these limitations by directly detecting stable planes from partial observations without the need for complete 3D object models. Unlike previous methods, the UOP method is more generalizable and adaptable to real-world scenarios with partial observations.

\noindent\textbf{Unseen object placement.} Previous studies on unseen object placement focused on the identification of stable placements that satisfy human preferences. For example, Jiang et al.\cite{jiang2012learninga} trained a classifier using a hand-crafted dataset to identify these placements; this approach relies on heuristic labels and requires complete observability. Cheng et al.\cite{cheng2021learning} proposed a deep learning model based on simulations to address the issue of heuristic labels; however, this approach was limited to task-specific objects. Another common approach\cite{mitash2020task} for placing unseen objects is using bounding box fitting to determine the shape and orientation of the object. This method can be fast and effective; however, it ignores the geometry of the object and relies only on its bounding box. Although this approach can be applied to unseen objects, it may not stably place objects in all situations, and therefore, it may be less effective than methods that consider the geometry of the object. In contrast, our approach can stably place unseen objects on a horizontal surface using only a single partial observation. Our method can handle a broad range of objects instead of being limited to specific object types.

\noindent\textbf{Robotic applications of object placements.} Prior works on object placement for robotic applications focused on solving specific tasks such as constrained placement\cite{mitash2020task}, upright placement\cite{newbury2021learning, pang2022upright}, and rearrangement\cite{wada2022reorientbot, paxton2022predicting}. However, these methods have several limitations. For example, Mitash et al.'s approach\cite{mitash2020task} relies on multi-view shape fitting and requires access to object models that may not be available in some scenarios. The deep learning approach proposed in \cite{paxton2022predicting} is limited to determining the required rotation for stable placements of objects in an upright orientation. Li et al.\cite{li2022stable} proposed a method that can only predict rotations that maintain objects in positions that maximize their height; these limitations restrict the applicability and potential of these methods for more general object placement tasks such as stacking and packing. In contrast, our approach addresses the fundamental problem of placing unseen objects on a horizontal surface and has the potential to be applied to a wider range of robotic applications.

\begin{figure*}[ht!]
    \centering
        \includegraphics[width=\textwidth]{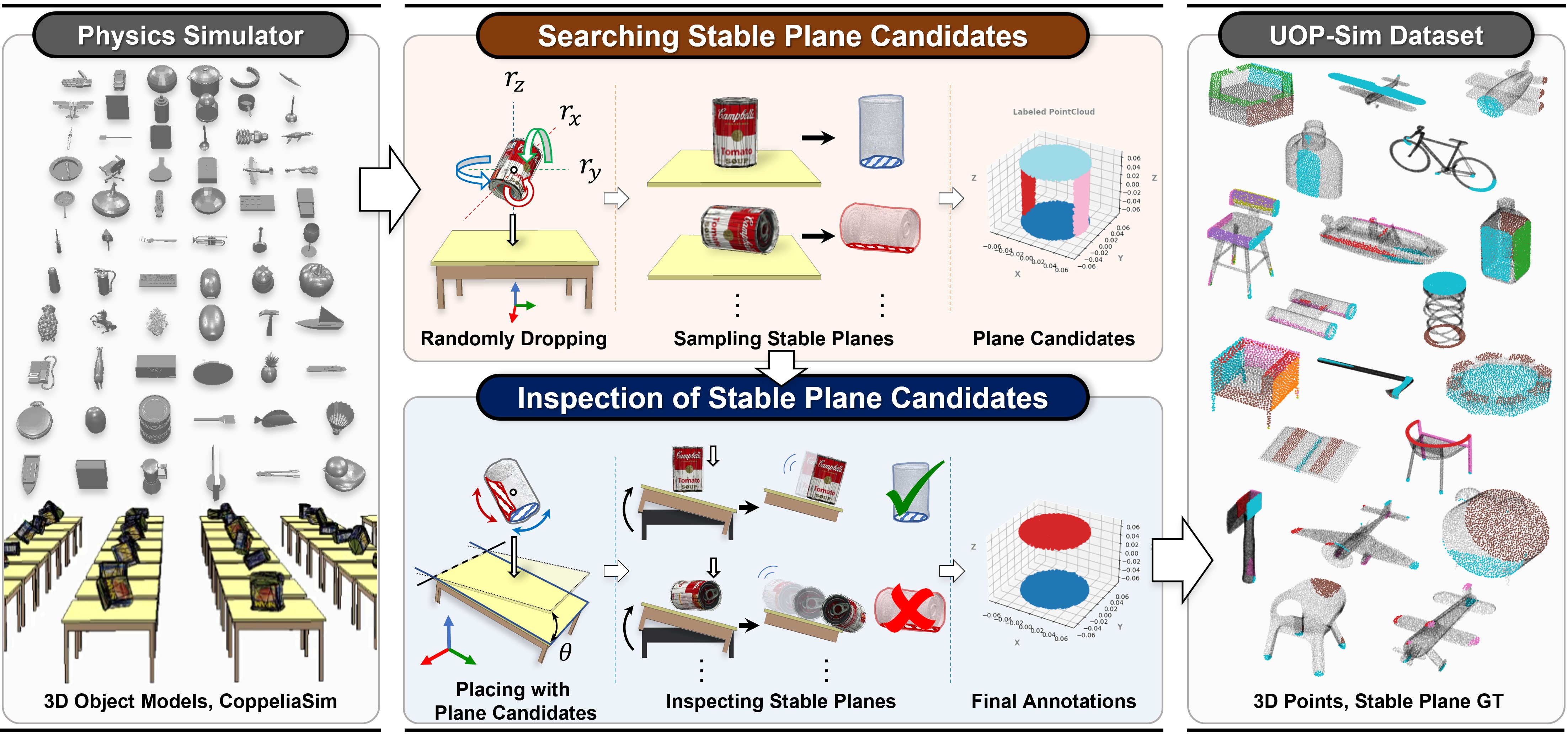}
  \caption{\textbf{UOP-Sim dataset generation pipeline.} The UOP-Sim dataset is a large-scale synthetic dataset that contains 3D object models (17.4K) and annotations of stable planes (69K). The dataset is generated by dropping each object on a table in 512 different configurations and by sampling stable planes that satisfy Eq.\ref{eq:stability}. The stable plane candidates are verified using a tilted table. }
    \label{fig:UOP-sim}
\end{figure*}

\section{Problem Statements}

\subsection{Assumptions}
The suggested approach is used when a robot must retrieve unseen objects that are mixed up in a container, or when a person hands a novel object to a robot and the robot does not know its correct orientation. The camera pose is assumed to be known in the workspace. The robot begins by grasping an object and capturing the scene using a single-view RGB-D camera. The resulting partial point cloud of the object is fed into the model to predict the most stable plane.

\subsection{Definitions}

\noindent\textbf{Point cloud}: Let ${X} \in \mathbb{R}^{N \times 3}$ be point clouds obtained by capturing the manipulating scene in which the robot grasps the object from the camera. 

\noindent\textbf{Object instability and stable planes}: Let $\mathcal{U}$ denote the instability of an object model. We define instability of an object model as the average of movements in a simulator over discrete time step $L$. Stable planes $\mathcal{S}$ are annotated for each object model that satisfies the condition $\mathcal{U} < \epsilon$. A stable plane $s \in \mathcal{S}$ is represented as normal vector $\vec{V}\in \mathbb{R}^3$, and threshold $\epsilon$ indicates that the object has stopped. 

\noindent\textbf{Dataset and deep learning model}: The dataset $\mathcal{D} = \{(\mathcal{O}, \mathcal{S})_n\}_{1}^{N}$ represents the $N$ set of object models $\mathcal{O}$ and corresponding stable planes $\mathcal{S}$ as the annotations. The function $\mathcal{F}: {X} \rightarrow s$ denote a deep learning-based model that considers point clouds ${X}$ as the input and produces the most stable plane $s$ as the output.

\noindent\textbf {Seen and unseen objects}: The set of object models used for training and testing the function $\mathcal{F}$ are denoted as $\mathcal{O}_{train}$, $\mathcal{O}_{test}$. If $\mathcal{O}_{train} \cap \mathcal{O}_{test} = \emptyset$, then objects $\mathcal{O}_{train}$ are considered seen objects while objects $\mathcal{O}_{test}$ are unseen objects for the model $\mathcal{F}$.

\subsection{Objectives} 
Our objective is to detect the most stable plane for placing unseen objects from a single-view observation. We aim to develop a function $\mathcal{F}: {X} \rightarrow s$ that minimizes the instability of the object $\mathcal{U}$.

\section{Learning for Unseen Object Placement}
We address the challenges of the UOP task, which is difficult to solve because of the need for a large-scale dataset for approximating stable planes and the complexity of the relationship between point clouds and annotated planes. We present a novel approach by introducing the UOP-Sim dataset to mitigate these challenges; this dataset includes 17K 3D object models and 69K labeled stable planes, and a UOP-Net neural network that can detect robust stable planes from partial point clouds. We propose a general and adaptable approach to the UOP task using these tools, which enables robots to accurately place unseen objects in real-world scenarios.

\subsection{UOP-Sim Dataset Generation}

\begin{figure*}[ht!]
   \centering
      \includegraphics[width=\textwidth]{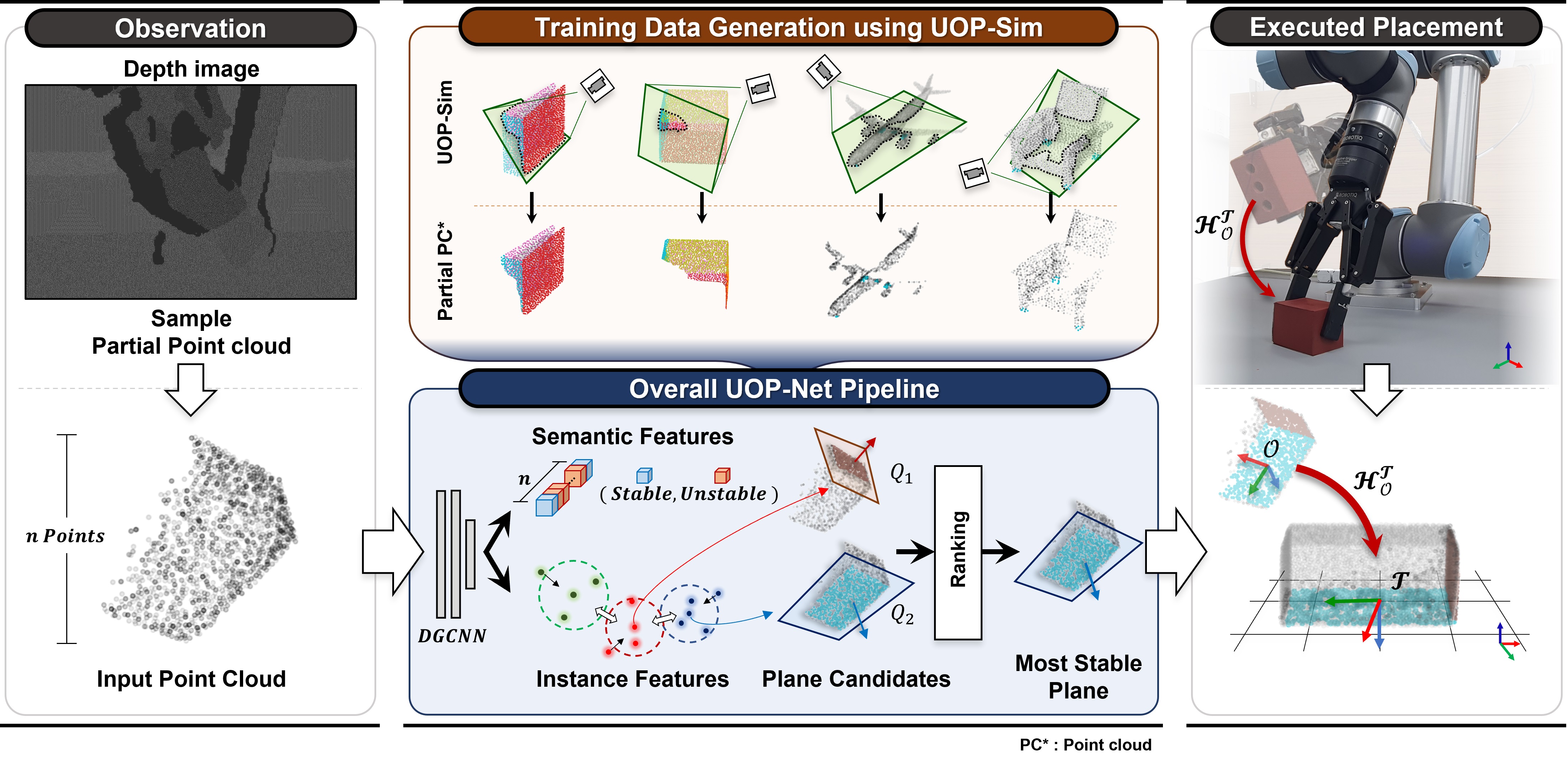}

    \caption{\textbf{Overall pipeline of UOP Method}. UOP detects the most stable plane directly from single-view and partial point cloud. UOP-Net is trained on the UOP-Sim dataset, and takes in a partial point cloud to predict the stable plane. The estimated stable plane is used to execute object placement based on the angle difference between the normal vector of the plane and the negative gravity vector.
    }

  \label{fig:UOP-net}
\end{figure*}

\noindent\textbf{Stable plane annotation.}
We defined the movement of the object at time step i as $\mathcal{M}_{i}$ in terms of its translation and rotation change in the world coordinates to evaluate the instability of the pose of the object in dynamic simulation. The pose can be represented as $\mathbf{H} =[\mathbf{R}|\mathbf{T}] \in \mathbb{SE}(3)$, where $\mathbf{R},\mathbf{T}$ are rotation and translation matrix. We tracked the pose of the object at each time step and calculated the difference between the consecutive poses (Eq.\ref{eq:movement}). Then, we took the average of these differences over a certain period ${L}$ to estimate the instability of the object at a time step ${i}$ (Eq.\ref{eq:stability}). To ensure robust annotation, we consider a range of discrete time step ${L}$ rather than only single time step ${i}$.

\begin{equation}
\label{eq:movement}
\mathcal{M}_i = ||\mathbf{H}_i- \mathbf{H}_{i-1}||_2
\end{equation}

\begin{equation}
\label{eq:stability}
\mathcal{U}_{i}=
\begin{cases}
{\frac{1}{L}}\sum_{j=i-L+1}^i\mathcal{M}_{j}, & \mbox{if } i >= L \\
{\frac{1}{i}}\sum_{j=1}^i\mathcal{M}_{j}, & \mbox{otherwise } 
\end{cases}
\end{equation}

We generated 512 orientations by dividing the roll, pitch, and yaw into eight intervals to explore a wide range of possible poses for the object. The object was then placed on a table with a random pose along the normal direction of the table. We dropped the object on the table and recorded all poses in which it remained stable (${\mathcal{U}_{i} < \epsilon_1}$) to identify stable planes that support the object.

We then used the density-based spatial clustering of applications with the noise algorithm\cite{ester1996density} to cluster the sampled poses. This allowed us to identify stable planes by clustering the poses along the z-axis; this represents the normal vector at the contact points of a stable plane with a horizontal surface. Subsequently, we masked the bottom 5\% of height along the surface normal of table to indicate areas that could support the stability of the object.

Specific planes may not be easily generalized because real-world environments cannot be perfectly simulated. This can be a problem for spherical model planes or the sides of a cylinder. We placed each object on a flat table with a normal vector of the plane candidates and tilted the table by 10° to address the issue. We then estimated the movement of the object across each time step and eliminated any planes that did not satisfy the condition $\mathcal{U}$ less than $\epsilon_2$. This allowed us to label stable planes that were robust for application to a horizontal surface, as shown in the samples of the UOP-Sim dataset in Fig.\ref{fig:UOP-sim}. The UOP-Sim contains a total of 17,408 3D object models and 69,027 stable plane annotations. Furthermore, our dataset contains both explicit and implicit planes, such as a flat surface formed by four chair legs. Supplementary Figure.1 contains additional sample images from UOP-Sim (YCB objects).

\noindent\textbf{Simulation environment setting}
We used PyRep\cite{james2019pyrep} and CoppeliaSim\cite{rohmer2013v} to build a simulation environment for computing object instability ($\mathcal{U}$). For the physics engine, we employed the bullet engine. Further, we used 3D object models from three benchmark datasets (3DNet\cite{wohlkinger20123dnet}, ShapeNet\cite{chang2015shapenet}, and YCB\cite{calli2015ycb}), which yielded a total of 17,408 models. We built 64 table models in the simulation environment to facilitate the annotation process.

\subsection{UOP-Net} 
\noindent\textbf{Network Architecture.} 
The UOP-Net is based on DGCNN\cite{wang2019dynamic} architecture and JSIS3D\cite{pham2019jsis3d} model. The network architecture includes three EdgeConv layers which are used to extract geometric features. These three EdgeConv layers use three shared fully-connected layers with sizes 64, 128, and 256. A shared fully-connected layer with size 1024 was then used to aggregate information from the previous layers. The global feature of point cloud was obtained using the Max-pooling operation, and two branches are used to transform the global features: one branch for semantic segmentation (which predicts whether a point is stable or unstable), and another branch for embedding instance features of stable planes. Both branches use fully-connected layers with sizes of 512 and 256. Before the two branches, LeakyReLU and batch normalization are applied to all layers.

A mean-shift clustering algorithm \cite{comaniciu2002mean} is applied to the predicted stable points for identifying the stable points, and RANSAC \cite{fischler1981random} is used to fit planes onto the clustered points\cite{Zhou2018}. Stability scores for each plane are calculated by the element-wise multiplication of semantic logits, predicted instance labels, and number of points composing each plane. Then, the plane with the highest score is output after fitting the planes and assigning stability scores based on the number of inliers that constitute the planes. The rotation matrix $\mathbf{R}$ is then determined by estimating the angular difference between the predicted normal vector of the stable plane and the gravity vector (negative table surface normal).

\subsection{Loss Function}
Our loss function comprise two terms; stability loss as $\mathcal{L}_{stability}$ and plane loss $\mathcal{L}_{plane}$,

\begin{equation}
\label{eq:the_sum_of_losses}
\mathcal{L} = \lambda_1 * \mathcal{L}_{stability} + \lambda_2 * \mathcal{L}_{plane},
\end{equation}

\noindent where $\lambda_1$ and $\lambda_2$ are hyper-parameters, setting as $\lambda_1=10$ and $\lambda_2=1$ respectively. 
The stability loss $\mathcal{L}_{stability}$ is defined by the binary cross-entropy loss to encourage predicted point label to match with the ground truth label. We adopted the discriminative function for 2D images\cite{de2017semantic} and 3D point cloud\cite{pham2019jsis3d} to embed instance features for the plane loss $\mathcal{L}_{plane}$.

\begin{table*}[ht!]
\caption{{UOP} Performance of UOP-Net and other baselines on the three benchmark objects (partial shape) in the simulation.}
\centering
\label{tab:simulation-partial}
\resizebox{\textwidth}{!}{%
{\renewcommand{\arraystretch}{1.2}
\LARGE{
\begin{tabular}{|cccccccccccccc|}
\hline
\multicolumn{2}{|c|}{\multirow{3}{*}{\begin{tabular}[c]{@{}c@{}}\\ Object type \& \\ Dataset\end{tabular}}} & \multicolumn{8}{c|}{ Object stability (S)} & \multicolumn{4}{c|}{\multirow{2}{*}{\begin{tabular}[c]{@{}c@{}}Success rate of \\ object placement (SR, $\%$)\end{tabular}}} \\ \cline{3-10}
\multicolumn{2}{|c|}{} & \multicolumn{4}{c|}{Rotation (R, $^{\circ}$) $\downarrow$} & \multicolumn{4}{c|}{Translation (T, $cm$) $\downarrow$} & \multicolumn{4}{c|}{} \\ \cline{3-14} 
\multicolumn{2}{|c|}{} & CHSA\cite{haustein2019object} & BBF\cite{mitash2020task} & RPF\cite{fischler1981random} & \multicolumn{1}{c|}{\begin{tabular}[c]{@{}c@{}}UOP \\ \textbf{(Ours)}\end{tabular}} & CHSA\cite{haustein2019object} & BBF\cite{mitash2020task} & RPF\cite{fischler1981random} & \multicolumn{1}{c|}{\begin{tabular}[c]{@{}c@{}}UOP \\ \textbf{(Ours)}\end{tabular}} & CHSA\cite{haustein2019object} & BBF\cite{mitash2020task} & RPF\cite{fischler1981random} & \begin{tabular}[c]{@{}c@{}}UOP \\ \textbf{(Ours)}\end{tabular} \\ \hline
\multicolumn{1}{|c|}{\multirow{5}{*}{\begin{tabular}[c]{@{}c@{}}Partial \\ Point cloud\end{tabular}}} & \multicolumn{1}{c|}{3DNet\cite{wohlkinger20123dnet}} 
& 28.81 & 33.10 & 25.40 & \multicolumn{1}{c|}{\textbf{8.68}} 
& 3.02 & 3.45 & 2.49 & \multicolumn{1}{c|}{\textbf{0.77}}  
& 52.60 & 36.12 & 64.34 & \textbf{65.24}  \\ 
\multicolumn{1}{|c|}{} & \multicolumn{1}{c|}{ShapeNet\cite{chang2015shapenet}} 
& 31.65 & 38.14 & 21.83 & \multicolumn{1}{c|}{\textbf{7.91}} 
& 3.97 & 4.53 & 2.46 & \multicolumn{1}{c|}{\textbf{0.86}} 
& 51.05 & 29.95 & 69.70 & \textbf{72.82}  \\ 
\multicolumn{1}{|c|}{} & \multicolumn{1}{c|}{YCB\cite{calli2015ycb}} 
& 34.85 & 41.69 & 22.41 & \multicolumn{1}{c|}{\textbf{5.92}} 
& 5.07 & 5.89 & 2.55 & \multicolumn{1}{c|}{\textbf{0.56}} 
& 42.48 & 30.41 & 62.13 & \textbf{73.32} \\ 
\hhline{|~|=|====|====|====|} 
\multicolumn{1}{|c|}{} & \multicolumn{1}{c|}{Total avg.} 
& 31.18 & 36.72 & 23.69 & \multicolumn{1}{c|}{\textbf{7.73}} 
& 3.82 & 4.39 & 2.50 & \multicolumn{1}{c|}{\textbf{0.74}} 
& 49.43 & 33.01 & 65.07 & \textbf{69.35} \\ \hline 
\end{tabular}}}}
\end{table*}


\section{Experiments}
\subsection{Comparison with Traditional Method in Simulation}
\noindent\textbf{Datasets.} 
We obtained a total of 152, 57, and 63 object categories in the 3DNet\cite{wohlkinger20123dnet}, ShapeNet\cite{chang2015shapenet}, and YCB\cite{calli2015ycb} datasets, respectively. We labeled the YCB object models in the simulation, but they were excluded from the training set to allow us to use the test set in both the simulation and real-world experiments. We excluded objects that had no stable planes (e.g., spherical objects) to ensure the quality of our dataset. Then, we splited the dataset into training and validation sets in a 8:2 ratio. The training set contained 13,926 objects and 55,261 annotations, while the validation set contained 3,482 objects and 13,766 annotations.

\noindent\textbf{Training Details.} We trained UOP-Net using partial point clouds sampled from the UOP-Sim dataset. During training, 2,048 points randomly sampled for each object and they underwent various types of data augmentation techniques such as rotation, sheering, point-wise jitter, and adding Gaussian noise to improve the performance of the model in real-world scenarios. The model was implemented using PyTorch\cite{paszke2019pytorch} and trained on an NVIDIA Titan RTX GPU with a batch size of 32 and a total of 1,000 epochs. We employed early stopping with a patience of 50 and used the Adam optimizer at a learning rate of 1e-3 to prevent overfitting.

\noindent\textbf{Baselines.} 
We compared the performance of our method with those of the following baselines:
\begin{itemize}
    \item \textbf{Convex hull stability analysis (CHSA)}\cite{haustein2019object, trimesh, hagelskjaer2019using}: The baseline method for determining stable object poses involves the calculation of the rotation matrix to allow an object to rest stably on a flat surface. The center of mass of the object is sampled, and the stable resting poses of the object on a flat surface are determined using the convex hull of the object. Then, the probabilities of the object landing in each pose were evaluated, and the pose with the highest probability was output.

    \item \textbf{Bounding box fitting (BBF)}\cite{mitash2020task, Zhou2018}: The method involves fitting an oriented bounding box to the convex hull of the object using principal component analysis (PCA) to minimize the difference between the volume of the convex hull and that of the bounding box. The object was placed on a planar workspace with the largest area facing down.
    
    \item \textbf{RANSAC plane fitting (RPF)}\cite{fischler1981random, Zhou2018}: The approach segments planes in a point cloud by fitting a model of the form $ax + by + cz + d = 0$ to each point $(x, y, z)$. Then, it samples several points randomly and uses them to construct a random plane while repeating this process iteratively to determine the plane that appears most frequently.
\end{itemize}

\begin{table}[ht!]
\caption{UOP Performance of UOP-Net and other baselines in the simulation (whole shape).The best and second-best results are indicated in \textbf{bold} and \underline{underline}, respectively.}
\label{tab:simulation-whole}
\tiny
\resizebox{0.48\textwidth}{!}{%
\begin{tabular}{cccccc}
\hline
\multicolumn{2}{c}{\multirow{2}{*}{\begin{tabular}[c]{@{}c@{}} \\Object type \&\\ Dataset\end{tabular}}} & \multicolumn{4}{c}{\begin{tabular}[c]{@{}c@{}}Success rate of \\ object placement (SR, \%)\end{tabular}} \\ \cline{3-6} 
\multicolumn{2}{c}{}                                                                                  
& CHSA\cite{haustein2019object}          & BBF\cite{mitash2020task}         & RPF\cite{fischler1981random}         & \begin{tabular}[c]{@{}c@{}}UOP-Net\\ (Ours)\end{tabular}         \\ \hline
\multirow{4}{*}{\begin{tabular}[c]{@{}c@{}}Whole\\ Point cloud\end{tabular}}       
& 3DNet\cite{wohlkinger20123dnet}   & \textbf{83.28}    & 55.48    & 70.89    & \underline{80.16}  \\ \cline{2-6}
& ShapeNet\cite{chang2015shapenet}  & \textbf{92.37}    & 49.09    & 79.00    & \underline{86.14}  \\ \cline{2-6}
& YCB\cite{calli2015ycb}            & \textbf{86.08}    & 45.86    & 78.11    & \underline{84.32}  \\ \cline{2-6}
& Total avg.                        & \textbf{86.31}    & 51.24    & 74.90    & \underline{82.79}  \\ \hline
\end{tabular}%
}
\end{table}

\noindent\textbf{Evaluation metrics.} 
We used two metrics to evaluate the performance of UOP-Net: object stability \textit{(OS)} and success rate of object placement \textit{(SR)}. We placed an object on a flat table and used the output of the model to estimate its stability for measuring \textit{OS}. We considered only rotational motion when we evaluated object stability because rotational motion is more common than translational motion when an object placed in an unstable state falls due to the vibrations. We evaluated the performance of the model by placing the object 100 times; we considered as a failure case if no planes were detected.

\begin{itemize}
    \item \textbf{Object Stability (OS)}: The metric quantifies the movement of the object during a discrete time step when it is placed on a horizontal surface using the predicted plane.
    \item \textbf{Success Rate (SR)}: The metric indicates the percentage of placements where the object remains stationary for a minute with accumulated rotation under $10^{\circ}$.
\end{itemize}

\begin{figure}[ht!]
    \centering
        \begin{subfigure}[t]{\columnwidth}
            \centering
            \includegraphics[width=\textwidth]{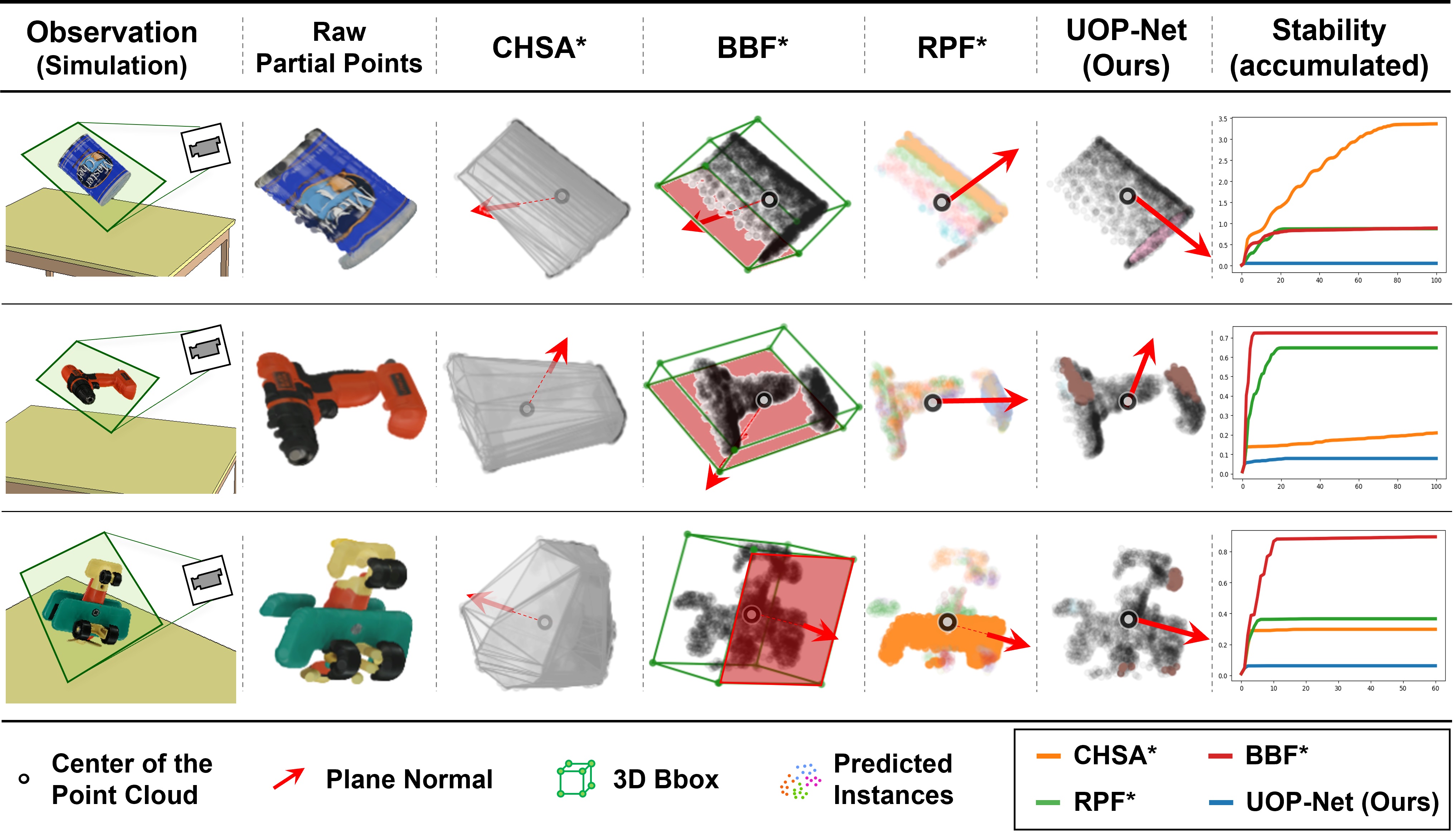}
        \end{subfigure}
        
    \caption{Visualization of simulation results on YCB\cite{calli2015ycb} objects (Red arrows are predicted normal vectors). The right graph depicts object stability; lower values are better.}
    \label{fig:sim_exp_result}
\end{figure}

\noindent\textbf{Discussion.} In the simulation experiments, we compared the performance of UOP-Net with the baseline methods on 3DNet\cite{wohlkinger20123dnet}, ShapeNet\cite{chang2015shapenet}, and YCB\cite{calli2015ycb} objects for partial shapes. Table \ref{tab:simulation-partial} presents the simulation results. Our UOP-Net achieved the best performance when single partial observations were used. We visualized the prediction results of each method and represented the object stability as a graph in Fig. \ref{fig:sim_exp_result}. The blue line in the graph (Fig. \ref{fig:sim_exp_result}) represents the case in which the object is placed in the plane predicted by the UOP-Net, and shows that the object can be placed more reliably compared with other methodologies.


When the object shape is fully visible, the CHSA method provides optimal results for stable object placements (Table \ref{tab:simulation-whole}). While the UOP-Net appears to be less effective in fully observed scenarios, the assumption is impractical. As shown in Table.\ref{tab:simulation-partial}, the UOP-Net method outperforms the CHSA method when only partial point clouds are available. This is because a convex hull derived from a partial point cloud lacks information about the invisible part of the object, thus making the determination of its center of mass more difficult. Additionally, we found that the CHSA method frequently predicts truncated planes. This is due to the truncated plane having the largest area within the convex hull, making it the most likely candidate for stable object placement (Fig.\ref{fig:sim_exp_result}).


The BBF method had the worst performance because it did not consider the geometric properties of the object, instead relying on placing the largest plane within the bounding box. This may work for objects with simple shapes; however, it is not suitable for complex shapes. Despite the better performance of the RPF method compared to BBF, it is still insufficient to identify consistently stable planes. Even though the RPF method eventually generates one of the most frequently sampled planes, its inability to select a reliably stable plane prevents successful object placement.  In contrast, the UOP-Net outstands at identifying the most stable plane even from partial observations, achieved by training the model to predict directly planes based solely on visible components.

Further, UOP-Net can detect explicit planes visible in the point cloud and implicit planes (e.g., a plane formed by the four legs of a chair) that are not necessarily visible but still contribute to the stability of the object. This is because our model was trained on a diverse set of objects and planes, and it can generalize new unseen objects and planes.


\subsection{Comparison with Learning-based Method in Simulation}

We compared the performance of UOP-Net, the Upright-Net \cite{pang2022upright} and the CHSA method alongside the point cloud completion method\cite{yu2021pointr}. To ensure a fair  comparison with Upright-Net, we conducted the experiments on a subset of UOP-Sim, where we used the same categorization scheme of Upright-Net. The dataset was divided into three splits: seen objects for training, seen objects for evaluation, and unseen objects for evaluation. Each category included 40 objects for training and 10 for evaluation. The categories are the following:

\begin{itemize}
    \item \textbf{\textit{Seen}}: Bed, Bench, Bottle, Bowl, Bus, Cabinet, Camera, Cap, Car, Chair, Jar, Laptop, Mug, Printer, and Table
    \item \textbf{\textit{Unseen}}: Basket, Helicopter, Lamp, Pot, Skateboard, Sofa and Tower
\end{itemize}

For the completion method, we used the Pointr\cite{yu2021pointr} that was trained on our subset of UOP-Sim; we confirmed that this model performs better than their pretrained model due to the difference in input preprocessing. Table \ref{tab:Additional Experiments} shows each method's object placement success rate in simulations. The rate is the ratio of successful placements to the total number of inference trials. Each object underwent 60 trials.

\begin{table}[ht!]

\centering
\caption{The placement success rates of UOP-Net, Upright-Net, CHSA with and without completion (C: 
 completion \cite{yu2021pointr})}
\label{tab:Additional Experiments}
\resizebox{0.48\textwidth}{!}{%
\begin{tabular}{cccccc}
\hline
\multirow{2}{*}{}        & \multirow{2}{*}{Data type} & \multicolumn{4}{c}{Method}            \\ \cline{3-6} 
                         &                            & CHSA w/o C & CHSA w/ C & Upright-Net\cite{pang2022upright} & UOP-Net (Ours) \\ \hline
\multirow{2}{*}{Whole}   & Seen                       & \textbf{92.27}          & -        & 77.97       & 87.59   \\ \cline{2-6} 
                         & Unseen                     & 86.76          & -        & 84.26       & \textbf{87.77}   \\ \hline
\multirow{2}{*}{Partial} & Seen                       & 57.04          & 67.06        & 17.04       & \textbf{76.75}   \\ \cline{2-6} 
                         & Unseen                     & 51.68          & 63.92        & 25.21       & \textbf{73.78}   \\ \hline
\end{tabular}%
}
\end{table}

\noindent\textbf{Comparison with Upright-Net}. Our proposed UOP-Net consistently outperformed Upright-Net \cite{pang2022upright} under all conditions. Upright-Net demonstrates comparable performance with the whole point cloud input; however, its performance was significantly lower than ours when the partial point cloud was input. This is due to Upright-Net's design, specifically constructed to predict only the upright orientation from the whole point cloud input. Consequently, it tends to fail when upright planes are invisible within the input partial point cloud. On the contrary, our UOP-Net was designed to detect all stable planes from the partial point cloud, thus leading to better performance. 

\begin{figure}[ht!]
    \centering
        \begin{subfigure}[t]{\columnwidth}
            \centering
            \includegraphics[width=0.8\textwidth]{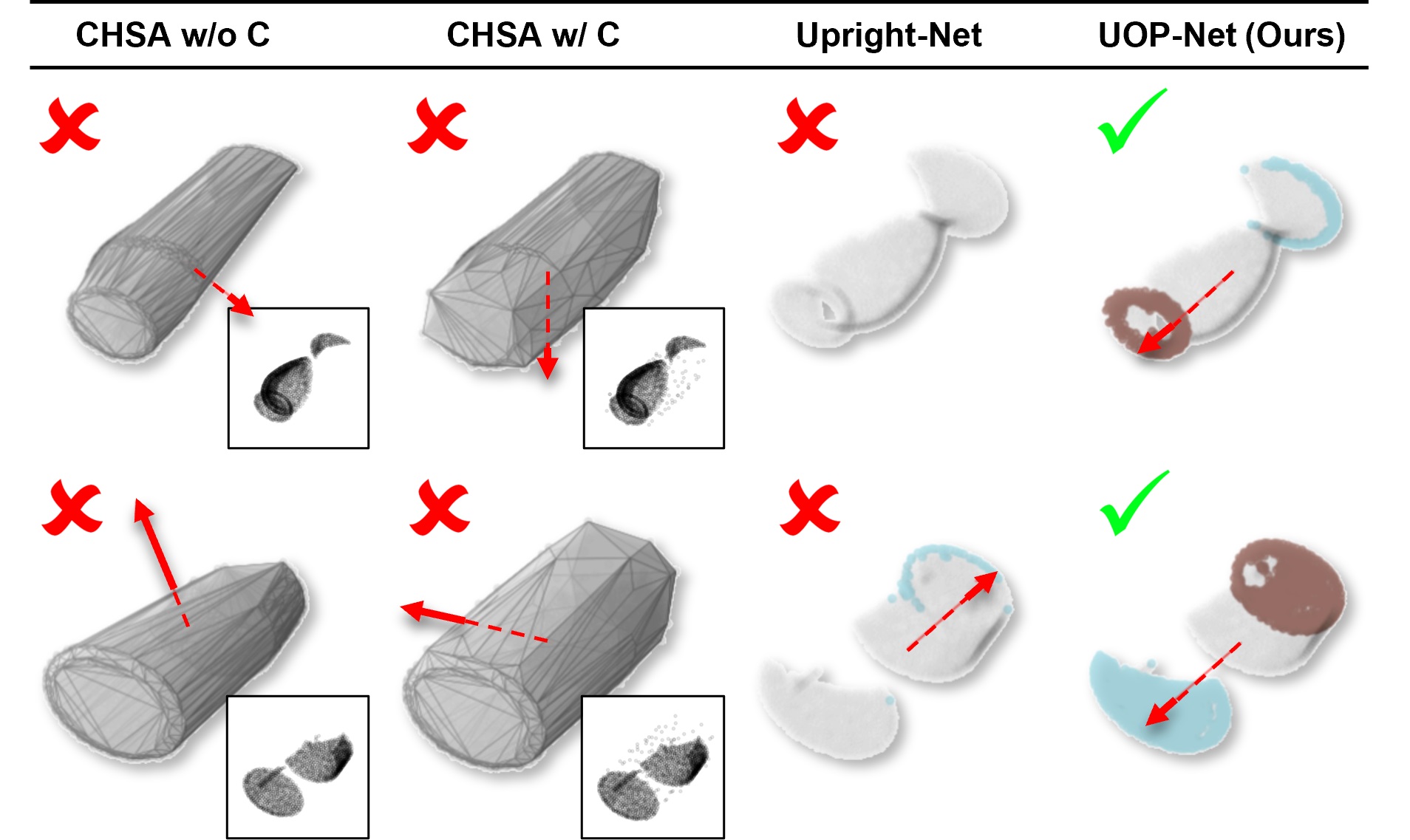}
        \end{subfigure}
        
    \caption{Comparison with learning-based method (only CHSA, CHSA + completion, Upright-Net and UOP-Net).}
    \label{fig:ablation_pred_res}
\end{figure}

\noindent\textbf{Comparison with CHSA + Completion}. Our UOP-Net outperforms CHSA in all cases except for seen objects provided with a complete point cloud. While the completion method does enhance CHSA's performance, it still tends to underperform compared to our method. This is primarily because the completion method often struggles to generate an accurate and detailed point cloud (Fig\ref{fig:ablation_pred_res}). 

\subsection{Robot Experiments}
We conducted real-world experiments to evaluate the performance of our UOP method. Our objectives were to demonstrate its ability to: (1) perform effectively in real-world scenarios despite being trained purely in simulation, (2) robustly handle partial and noisy observations, and (3) detect stable planes in unseen objects, achieving SOTA performance

\begin{figure}[ht!]
    \centering
        \begin{subfigure}[t]{\columnwidth}
            \centering
            \includegraphics[width=\textwidth]{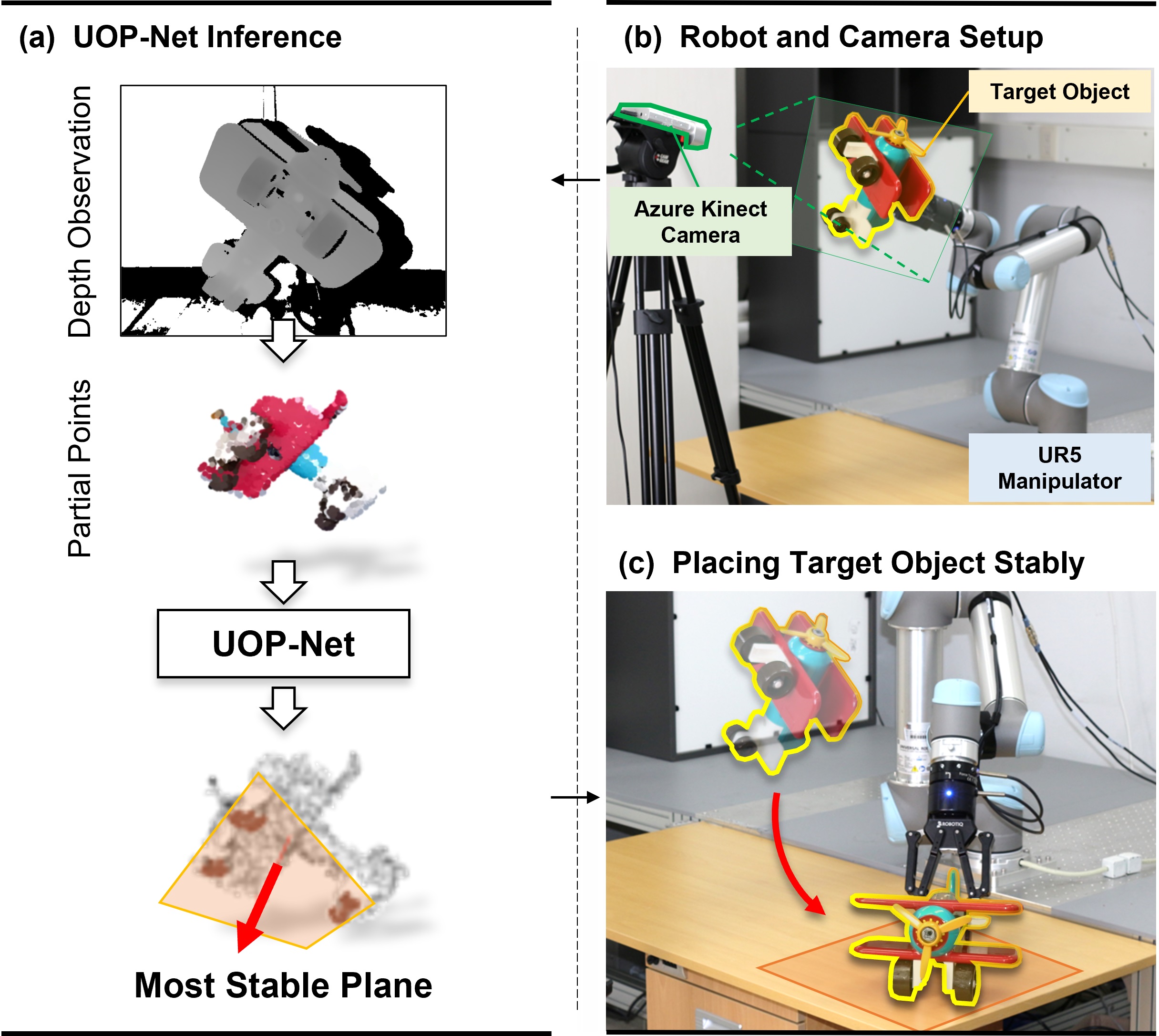}
        \end{subfigure}
        
    \caption{\textbf{Unseen object placement using UOP-Net}. (a) Given a segmented target object, UOP-Net considers the 3D point clouds observed by a depth camera (Azure Kinect) as inputs and predicts the most stable plane. (b) UR5 places the target object on the table.}
    \label{fig:real_experiment}
\end{figure}

\noindent\textbf{Real environment setting.}
We conducted experiments using a universal robot (UR5) manipulator and an Azure Kinect RGB-D camera to evaluate the performance of our object placement method in a real-world scenario (Fig.\ref{fig:real_experiment} (b)). We used the MANet \cite{fan2020ma} object segmentation method with a DenseNet121 \cite{iandola2014densenet} backbone to segment the target object and the gripper. We segmented the visible region of the target object from the RGB image and cropped the depth image using a mask. Then, we sampled the point cloud from the depth image using voxel-down sampling \cite{Zhou2018} and fed it to UOP-Net. The model predicted the most stable plane and calculated the rotation value between the plane and the table. Then, the UR5 robot placed the target object on the table. We utilized the BiRRT algorithm \cite{qureshi2015intelligent} implemented with PyBullet \cite{coumans2021} and integrated it with collision checking in a physics engine to ensure smooth planar motion.

\noindent\textbf{Evaluation metrics.} To ensure a fair comparison, we strived to standardize the object grasp configurations across various methodologies. Subsequently, we adopted the SR metric to quantitatively assess the efficacy of our proposed approach in real-world scenarios. Throughout the experiments, we instructed the robotic system to execute object placements onto the surfaces predicted by the UOP-Net model. The successful placement was predicated based on visual confirmation of the object maintaining a stable, non-sliding position on the predicted plane. Conversely, if the model failed to identify any viable stable planes, the trial was classified as unsuccessful. For each distinct object, we conducted a series of 10 placement trials to ensure a comprehensive evaluation outcome.

\begin{table}[ht!]
\centering
\caption{UOP performance of the UOP-Net and baselines on YCB\cite{calli2015ycb} in the real world.}
\label{tab:real world-eval}
\tiny
\resizebox{0.48\textwidth}{!}{%
\begin{tabular}{ccccc}
\hline
YCB & CHSA\cite{haustein2019object}  & BBF\cite{mitash2020task}  & RPF\cite{fischler1981random} & \begin{tabular}[c]{@{}c@{}}UOP-Net\\ (Ours)\end{tabular} \\ \hline
Coffee can & 0 & 0 & 4 & \textbf{10} \\
Timer & 0 & 1 & \textbf{6} & \textbf{6} \\
Power drill & 0 & 1 & \textbf{5} & \textbf{5} \\
Wood block & 1 & 1 & \textbf{10} & \textbf{10} \\
Metal mug & 0 & 0 & 6 & \textbf{9} \\
Metal bowl & \textbf{10} & 5 & \textbf{10} & \textbf{10} \\
Bleach cleanser & 3 & 3 & \textbf{9} & \textbf{9} \\
Mustard container & 2 & 0 & 5 & \textbf{10} \\
Ariplane toy & 0 & 3 & 0 & \textbf{4} \\
Sugar box & 2 & 3 & \textbf{10} & \textbf{10} \\
Chips can & 2 & 0 & 8 & \textbf{10} \\
Banana & 5 & 5 & \textbf{9} & \textbf{9} \\ \hline
Average & 2.1 & 1.8 & 6.8 & \textbf{8.5} \\ \hline
\end{tabular}%
}
\end{table}

\begin{figure}[ht!]
    \centering
        \begin{subfigure}[t]{\columnwidth}
            \centering
            \includegraphics[width=\textwidth]{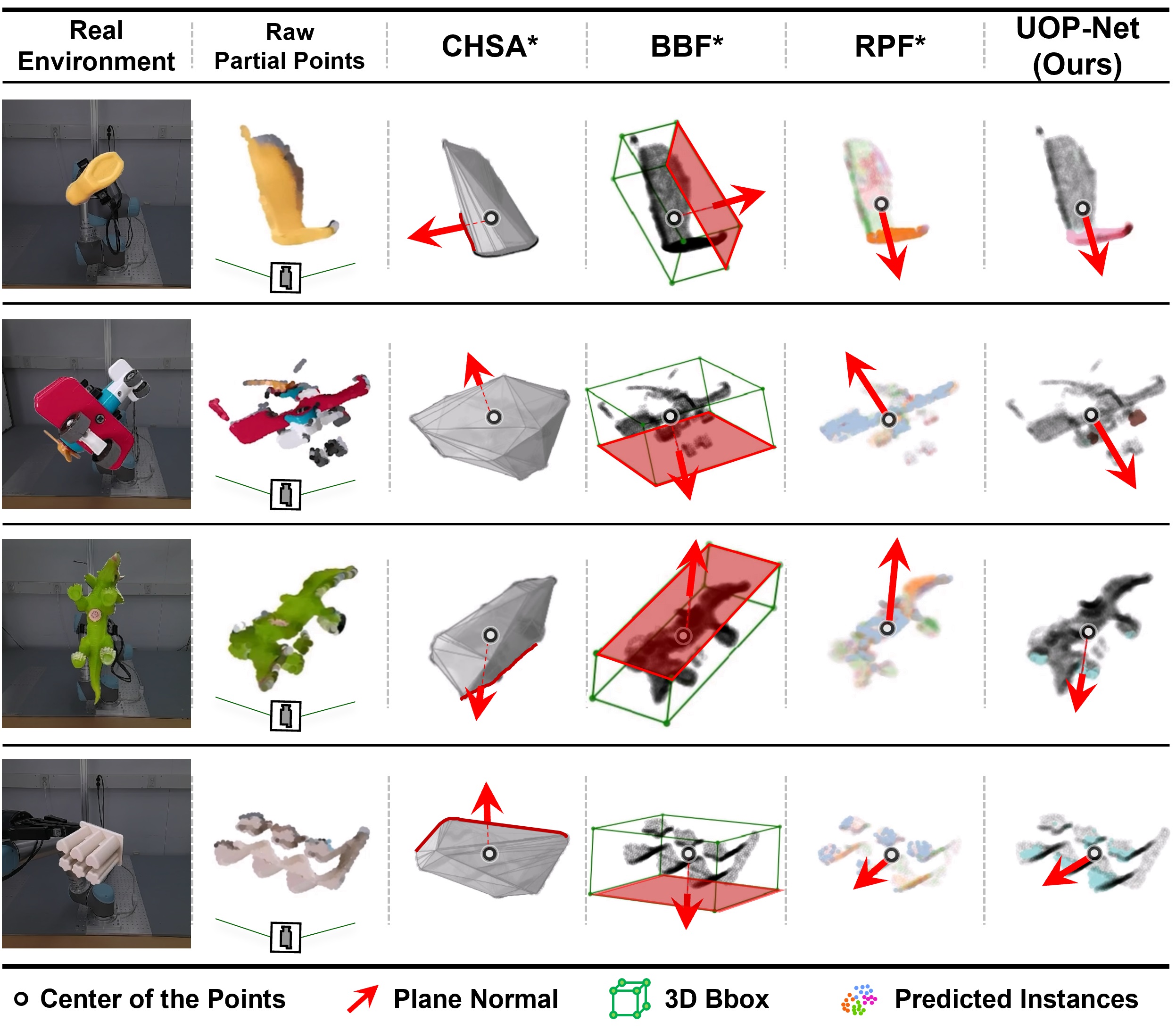}
        \end{subfigure}
        
    \caption{Visualization of the real world results. (rows 1, 2: YCB objects, rows 3, 4 : novel objects)}
    \label{fig:comparison_pred_res}
\end{figure}

\noindent\textbf{Results.}
We selected 12 objects from the YCB dataset. Objects with spherical shapes (e.g., apples), dimensions that were extremely small, or small depth values were excluded from the test set. Table \ref{tab:real world-eval} indicates that our method outperforms other baselines in terms of the success rate across all objects. Although real-world perceptions are noisy, UOP-Net provides a stable plane that can be attributed to our model learning from partial point clouds captured by a depth camera and corrupted by noise. Other benchmarks (CHSA and primitive shape fitting) performed extremely poorly because they could not obtain the complete shape of the object in the real world and were unable to respond to sensor noise. We evaluated our method on completely new objects that did not have an available CAD model (a dinosaur figurine and an ice tray, as shown in Fig.\ref{fig:comparison_pred_res}) to verify further that our model can perform on unseen objects. UOP-Net detected implicit planes (e.g., the four legs of the dinosaur) even though the object shapes were complex.

\section{Conclusion} 
In this study, we presented UOP-Net, a novel method developed to detect stable planes of unseen objects. We also introduced an approach to annotate  automatically stable planes for various objects, and the large-scale synthetic dataset, called UOP-Sim, was generated. Our dataset contains 17.4K 3D objects and 69K stable plane annotations. The effectiveness of UOP-Net was demonstrated by achieving SOTA performance on objects from three benchmark datasets, thus indicating its accuracy and reliability in detecting stable planes from unseen and partially observable objects.

\section*{Acknowledgment}
\begin{spacing}{0.4}
{\scriptsize This research was completely supported by a Korea Institute for Advancement of Technology (KIAT) grant funded by the Korea Government (MOTIE) (Project Name: Shared autonomy based on deep reinforcement learning for responding intelligently to unfixed environments such as robotic assembly tasks, Project Number: 20008613). This research was partially supported by an HPC Support project of the Korea Ministry of Science and ICT and NIPA.}
\end{spacing}

\bibliography{references.bib}
\bibliographystyle{IEEEtran}

\end{document}